# Hybrid PS-V Technique: A Novel Sensor Fusion Approach for Fast Mobile Eye-Tracking with Sensor-Shift Aware Correction

Ioannis Rigas, Hayes Raffle, and Oleg V. Komogortsev

*Abstract*—**This paper introduces and evaluates a hybrid technique that fuses efficiently the eye-tracking principles of photosensor oculography (PSOG) and video oculography (VOG). The main concept of this novel approach is to use a few fast and power-economic photosensors as the core mechanism for performing high speed eye-tracking, whereas in parallel, use a video sensor operating at low sampling-rate (snapshot mode) to perform dead-reckoning error correction when sensor movements occur. In order to evaluate the proposed method, we simulate the functional components of the technique and present our results in experimental scenarios involving various combinations of horizontal and vertical eye and sensor movements. Our evaluation shows that the developed technique can be used to provide robustness to sensor shifts that otherwise could induce error larger than 5°. Our analysis suggests that the technique can potentially enable high speed eye-tracking at low power profiles, making it suitable to be used in emerging head-mounted devices, e.g. AR/VR headsets.**

## I. INTRODUCTION

Eye-tracking is expected to become an essential tool for seamless human-computer interaction (HCI) in modern head-mounted devices. For example, in the case of AR/VR headsets, eye-tracking can substantially improve the immersion and the overall user experience by enabling applications like foveated rendering [1], saccade-contingent screen updating [2], touchless interaction [3], and assist on the prevention of eye fatigue [4] and cybersickness [5]. In order to meet the demands of the growing mobile AR/VR ecosystems, two very important requirements for eye-tracking systems aiming to enable such applications are high tracking speed and relatively low power consumption.

I. Rigas and O.V. Komogortsev are with Texas State University, Department of Computer Science, 601 University Dr, San Marcos, TX 78666 USA, (e-mails: rigas.ioann@gmail.com; ok@txstate.edu).

H. Raffle is with Google, 1600 Amphitheater Drive, Mountain View, CA 94043, USA, (e-mail: hraffle@google.com).

Most current eye-tracking systems are based on the principle of video oculography (VOG). In a typical VOG implementation [6], the eye is illuminated by one (or more) infrared LED(s), and consecutive images of the eye are captured and processed to extract important features, e.g. pupil center and corneal reflection. The differences in position of these features can be used to estimate eye movement with relative robustness to small sensor movements. The systems based on VOG can provide high accuracy during gaze estimation but they have certain limitations when high speed eye-tracking is needed combined to low power consumption. These limitations arise from the need to capture and process multiple images, a procedure with considerable burden for computational resources. For binocular eye-tracking these demands and the overall cost become further inflated.

A number of alternative eye-tracking methods have been explored in the past, with the most prominent being: a) the magnetic scleral coil method [7], b) electrooculography (EOG) [8], and c) photosensor oculography (PSOG) [9]. Among them, PSOG appears to fulfill many of the eye-tracking needs posed by modern headsets. The principle of PSOG is based on the direct measurement of the amount of reflected light from the eye using simple pairs of photo-sensitive sensors. A major advantage of PSOG when compared to VOG is the minimal computational burden (just a few computations to combine sensor outputs) that can enable eye-tracking with high sampling rate and low power consumption. Also, the PSOG does not need any attachment on the eye or skin making it less obtrusive than the magnetic scleral coil method and EOG. Despite these obvious advantages, PSOG has also its Achilles' heel: it is very sensitive to sensor shifts. Most headsets use head-straps to limit excessive mobility, however, small sensor movements can still occur due to facial expressions or body movements (e.g. during jumping, walking). Such sensor shifts can result in considerable degradation of accuracy for the traditional implementations of PSOG.

In this work, we propose a new approach for addressing the limitations of traditional photosensor oculography and video oculography systems by selectively combining the best characteristics from both worlds. The key contributions of this work are:

1) We introduce the hybrid PS-V technique, a new eye-tracking concept based on the fusion of photosensor and video oculography principles. We present the details of the technique and simulate its functional components.

2) We perform an evaluation of the technique using the developed simulated framework. We feed the models with real eye movements and explore the baseline potential of the technique and the achieved robustness to sensor shifts.

## II. BACKGROUND

Eye-tracking techniques based on the direct measurement of the amount of reflected light from the eye have been investigated since the early 1950's [10]. Most of these techniques use invisible infrared light and rely on the existing differences in the reflectance properties of different regions of the eye (sclera, iris, and pupil). When the eye moves, the transitions between these regions can be tracked using simple pairs of photosensors positioned in close proximity to the eye. The term photosensor oculography (PSOG) can be used to collectively refer to the techniques based on this principle of operation but other alternative terms have been also used in the past, such as: photoelectric technique, infrared oculography, and limbus reflection method. Most of the PSOG techniques are based on the differential operation principle, i.e. they calculate relative differences between sensor pairs. In order to avoid ambient light interferences, systems based on PSOG can use modulated (chopped) light, as proposed in [11]. The PSOG techniques allow measurement of eye-ball rotations with very good precision (a few arc min), and additionally, the fast switching times of the sensors (usually at the order of ns) and the minimal computational complexity can enable tracking with high speed. However, in order to provide acceptable eye-tracking accuracy, a system based on PSOG needs to be firmly affixed to the head because even the slightest sensor shifts can induce large errors during gaze estimation. For example, sensor movements larger than 0.5 mm can result in accuracy error larger than 1°. A compact overview of the characteristics of various PSOG systems can be found in [9]. It should be noticed that even though several PSOG variations have been developed in order to advance the characteristics of this technology in terms of linearity, crosstalk, sensor placement, and tracking range [12-15], the lack of robustness during sensor shifts hindered the widespread adoption of PSOG. On the other hand, the breadth of technical advancement in recent years has focused on VOG techniques where algorithms have been developed to accommodate for such sensor shifts, giving the technology robustness in real-world conditions.

## III. THE HYBRID PS-V TECHNIQUE

### A. General Overview

The main goal of the hybrid PS-V technique is to address the sensor-shift related issues of PSOG while keeping the inherent advantages of this technology, such as the high speed and low power consumption. The developed approach to achieve this goal is based on the combination of information coming from two subsystems, a subsystem based on PSOG and a subsystem based VOG. The PSOG subsystem is used to track eye rotations at high sampling rate (1000 Hz or more). The VOG subsystem is used as the means for estimating sensor shifts and thus it can operate at a much lower sampling rate (e.g. 5 Hz or less). A basic assumption of the technique is the existence of a rigid connection between the PSOG and VOG subsystems, so that the sensor movement estimated by the VOG subsystem can be used to rectify the movement-induced artifacts appearing in PSOG subsystem. The rigid connection requirement is by-design fulfilled when the two subsystems are embedded in a headset setup. In Fig. 1 we present a summarizing overview of the functional components of the hybrid PS-V technique.

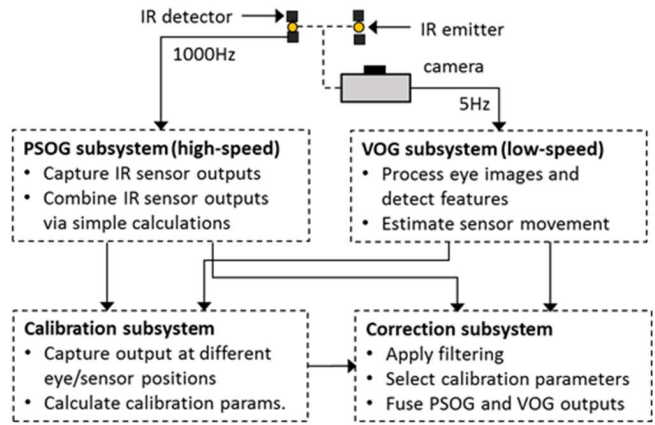

Fig. 1. Overview of the proposed hybrid PS-V technique.

The description and evaluation of the hybrid PS-V technique is performed using a semi-simulated framework. We use model-driven simulation to represent the functional components of the technique, and then, during the evaluation phase we use signals from real eye movements as input to the simulated models. While the use of real system components is essential for validating the final robustness of the technique, the initial evaluation of the proposed novel scheme via simulation can greatly facilitate the exploration of some very crucial aspects, such as: a) during development, it allows for the in-depth investigation and modeling of PSOG subsystem's behavior in the case of sensor movements, and b) during evaluation, it provides better control of the performed sensor movements (exact magnitudes and directions) and thus enables the assessment of sensor-shift robustness against a well-defined ground truth.

### B. Generation of 3-D Rendered Synthetic Eye Images

The first step in the developed simulation framework is to generate synthetic eye images using the simulation software introduced by Swirski and Dodgson in [16]. This software was built using Python and the 3-D graphics software Blender [17], and it includes realistic models of the human eye (supporting movements for eye, eyelid and pupil), the camera module, and the light sources. The 3-D scene elements can be algorithmically positioned and rotated (3-DoF) in order to simulate different eye-tracking scenarios. In our case, we use this software to generate the frames that are then used during the simulation of PSOG and VOG subsystems. To generate the frames used during PSOG subsystem's simulation we use two point-light sources (simulating the IR emitters) positioned

±1.4 cm horizontally, 1 cm under and 3 cm away from the pupil center (all distances are measured with respect to the eye in neutral position and refer to the left eye). The camera module is centered (horizontally and vertically) and 5 cm away from the pupil center with the field of view set to fully cover the eye area (in our case 45°). To generate the frames used during VOG subsystem's simulation we use exactly the same positioning for the point-light sources to simulate the simultaneous operation of the two subsystems (PSOG and VOG). This time, though, the camera module is placed 1 cm under and 5 cm away from the pupil center. In all cases, the resolution of the rendered frames is set to be 240 x 320 pixels. To rotate the simulated 3-D eye model, we either send pre-defined values of specific eye positions (during calibration) or send values recorded from real eye movements (during evaluation) using a high-grade eye-tracker as ground truth. To simulate sensor movements, we use pre-defined (ground truth) values to translate the camera module and the lights together, thus conforming with the rigid connection assumption.

## C. PSOG Subsystem

An infrared detector can be modeled as a controlled current source connected in parallel to an exponential diode [18]. The currents of the controlled source $I_p$ and the exponential diode $I_d$ can be modeled using Eq. 1-2:

$$I_p = R_\lambda \cdot P \qquad (1)$$

$$I_d = I_s \cdot \left( e^{\frac{q \cdot V_A}{k_B \cdot T}} - 1 \right) \qquad (2)$$

where $R_\lambda$ is the responsivity at wavelength $\lambda$, $P$ is the incident light power, $I_s$ is the reverse saturation current, $q$ is the electron charge, $V_A$ is the applied bias voltage, $k_B = 1.38 \cdot 10^{-2}\ J/K$ is the Boltzmann constant, and T is the absolute temperature. Furthermore, when operating in photovoltaic mode the photodiode is zero-biased (V = 0), and since $I_d \to 0$ the output of the sensor is analogous to the incident light power $P$ ($R_\lambda$ can be considered constant for given conditions).

In order to simulate the incident light power on the sensor after the light is reflected by the eye, we use a Gaussian modulated window binning operation applied on the pixel intensity values of selected areas of the 3-D rendered eye frames. The window $W_{i,j}$ is selected to be 13° x 13° and it is multiplied with a Gaussian kernel $G_{i,j}$ of σ = ¼ the window side. This operation results in the simulation of the receptive area of a photodiode with a half reception angle of about ±8°. The output of this photodiode is calculated by averaging the Gaussian-modulated pixel intensity values within the defined window, as shown in Eq. 3:

$$I_{PD} = \frac{\sum_{i,j} G_{i,j} \cdot W_{i,j}}{i \cdot j}, \ i, j = pixel\ coord. \qquad (3)$$

In Fig. 2, we show a graphical presentation of the steps followed during the simulation of a single photodiode's output, using an example frame from our experiments.

A PSOG setup usually contains two or more photodiodes positioned to capture the light reflected from different regions of the eye. Combining clues from previously proposed PSOG design principles [9, 12] we develop a setup based on two wide-angle emitters (simulated by the point-light sources) and four photodiodes (each simulated as described previously) positioned to split the eye region into four semi-overlapping detection quadrants. The diagram of the developed PSOG design and the respective detection areas are shown in Fig. 3. To calculate the horizontal/vertical components of eye movement ($I_{PSOG}^{H,V}$) with the developed design, we need to perform the low-complexity operations described in Eq. 4-5:

$$I_{PSOG}^H = \left( I_{PD_1} + I_{PD_4} \right) - \left( I_{PD_2} + I_{PD_3} \right) \qquad (4)$$

$$I_{PSOG}^V = \left( I_{PD_1} + I_{PD_2} \right) - \left( I_{PD_3} + I_{PD_4} \right) \qquad (5)$$

The developed PSOG design can be relatively flexible during a practical implementation since it can be alternatively realized using four narrow half-angle emitters paired with four photodiodes of wider half-angle. Also, to avoid occlusion of visibility the sensors can be placed more distant and slightly angled to point at the eye target areas.

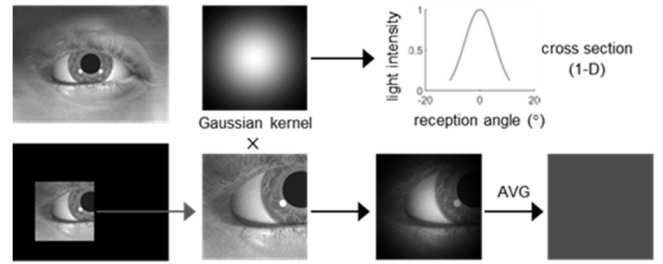

Fig. 2. Graphical presentation of the steps for the simulation of a single photodiode's output.

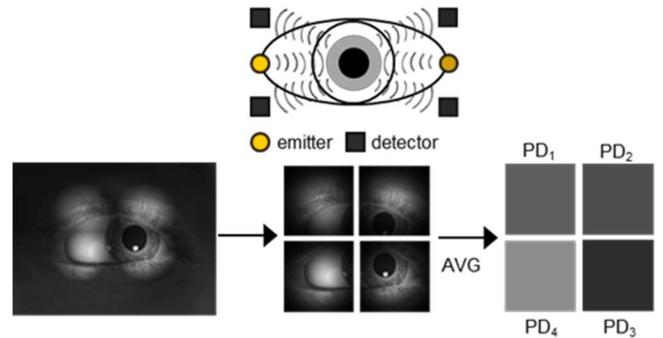

Fig. 3. The used PSOG design and the respective simulated detection areas.

## D. VOG Subsystem

The simulation of the output from the video sensor is more straightforward since the employed 3-D rendering software already provides a fully functional model of a camera sensor. Hence, the rendered frames from the VOG setup simulation can be used directly to represent the output of the video sensor. During our experiments the output of the video sensor is sampled at low rate (5 Hz) to accurately represent the required specifications for camera operation. The VOG subsystem further processes the frames in order to extract features and estimate the sensor movement that will be used to perform the correction of PSOG subsystem's output.

The methodology used for the estimation of sensor movements via the VOG subsystem is based on the quantification of the differences in relative movements of the pupil center (attached to the eyeball) and the corneal reflection, when eye and sensor movements occur. The basic principles of this methodology were investigated in [19] for the task of sensor movement compensation when performing eye-tracking using a pure VOG system. In our case, the VOG subsystem is used only as the mechanism for sensor movement estimation, and not for performing the full eye-tracking queue. For this reason, we focus only on the part related to the calculation of the camera (sensor) movement vector. Let us assume that we have available at each time the tracked positions of the pupil center $PC_{tr}$ and corneal reflection $CR_{tr}$. If we accept an approximately linear relationship for the relative movements of pupil center and corneal reflection when eye and sensor movements occur, then, we can use the formulas of Eq. 6-7 to separate the part of the apparent pupil movement that is generated exclusively from sensor movements ($PC_s^{H,V}$). For horizontal and vertical sensor movements we can use the parameters of the simulated VOG setup to convert this movement from pixel-space to millimeters of estimated sensor movement $x_{VOG}^{H,V}$.

$$PC_s^{H,V} = \frac{CR_{tr}^{H,V} - P_{tr}^{H,V} \cdot G_e^{H,V}}{G_s^{H,V} - G_e^{H,V}}$$ (6)

$$G_e^{H,V} = \frac{\Delta CR_e^{H,V}}{\Delta PC_e^{H,V}}, G_s^{H,V} = \frac{\Delta C_s^{H,V}}{\Delta PC_s^{H,V}}$$ (7)

The terms $G_e^{H,V}$ and $G_s^{H,V}$ represent the average eye and sensor movement gains, i.e. the fraction of the respective corneal reflection movement per unit of pupil center movement when each type of movement (eye or sensor) is performed separately. To find the values of $G_e^{H,V}$ and $G_s^{H,V}$ we simulated eye and sensor movements and calculated the average values of 0.38/0.39 and 0.83/0.81 respectively.

As we mentioned, to use Eq. 6-7 we need to have available at each time the tracked positions of the pupil center and corneal reflection. To track these positions we fed the 3-D rendered eye frames to the open-source tracking software Haytham [20]. The software allows to parameterize the thresholds for detecting pupil center and the closest (to it) corneal reflection. In our case (two light sources), this interchangeable detection of the closest corneal reflection makes possible the coverage of larger range of movements. It should be emphasized that the ambiguity regarding which corneal reflection is captured every time does not affect sensor movement estimation since Eq. 6-7 make use of relative differences expressed with reference to the primary eye position.

*E. Calibration Subsystem*

The calibration subsystem provides the composite mapping function for performing the tasks of: a) eye movement calibration, i.e. the transformation of the PSOG subsystem's output from raw units to degrees of visual angle, and b) sensor movement calibration, i.e. the transformation of calibration parameters in relation to sensor movement. After the calibration parameters are computed, they can be stored so that the correction subsystem can use them routinely to combine the outputs from PSOG and VOG subsystems, and generate the corrected output signal. To develop a powerful model for the calibration function, we first investigate the behavior of the raw output of the PSOG subsystem. The developed simulation framework greatly facilitates such a task because it allows to perform a controlled dense scan of eye and sensor positions and observe the general behavior of the output. In our case, the scan covers eye positions in range ±10° (horizontal/vertical) with step of 0.5°, and sensor positions in range ±2 mm (horizontal/vertical) with step of 0.5 mm. In Fig. 4, we show the respective clusters of curves that represent the general behavior of PSOG subsystem's output when sensor movements occur. We can observe that for the used PSOG design the sensor movements mainly affect the capturing of eye movements of the same direction. For example, horizontal sensor movements induce a significant translation of the horizontal output curves, whereas the vertical output curves are only slightly affected. An analogous effect can be observed for the case of vertical sensor movements. Also, we can see that the linearity is very good at the primary sensor position (0 mm) but gradually deteriorates as we move the sensor.

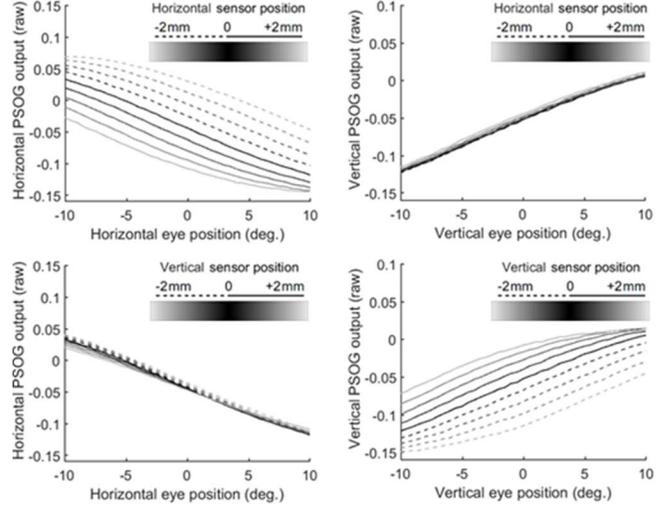

Fig. 4. Behavior of PSOG subsystem's output for different combinations of horizontal and vertical eye/sensor movements.

Based on the observed behavior we developed the calibration model of quadratic mapping functions described in Eq. 8-11. The calibration function $f_C^{H,V}$ is used to provide the mapping between pre-defined eye positions $x_e^{H,V}$ and the raw eye-tracking output from the PSOG subsystem ($I_{PSOG}^{H,V}$). The mapping is done via the top-level parameters $a^{H,V}$, $b^{H,V}$, $c^{H,V}$, which in turn are mapped at a lower-level as functions of sensor position. The training of calibration function is performed using data captured at a number of predefined eye positions $x_e^{H,V}$ (minimum 3/direction) and sensor positions $x_s^{H,V}$ (minimum 3/direction). Data fitting is done using Least Squares (LS) regression.

$$f_C^{H,V}(x_e^{H,V}, x_s^{H,V}) = a^{H,V}(x_s^{H,V}) \cdot x_e^{H,V^2} + b^{H,V}(x_s^{H,V}) \cdot x_e^{H,V} + c^{H,V}(x_s^{H,V})$$ (8)

$$a^{H,V}(x_s^{H,V}) = a_1^{H,V} \cdot x_s^{H,V^2} + a_2^{H,V} \cdot x_s^{H,V} + a_3^{H,V}$$ (9)

$$b^{H,V}(x_s^{H,V}) = b_1^{H,V} \cdot x_s^{H,V^2} + b_2^{H,V} \cdot x_s^{H,V} + b_3^{H,V}$$ (10)

$$c^{H,V}(x_s^{H,V}) = c_1^{H,V} \cdot x_s^{H,V^2} + c_2^{H,V} \cdot x_s^{H,V} + c_3^{H,V}$$ (11)

In Fig. 5, we show the fitted top-level (Eq. 8) calibration curves when using eye positions -10°, 0° and +10° and sensor positions -2, 0, +2 mm. The calibration curves are shown superimposed on the simulated eye-tracking curves previously shown in Fig. 4 (we focus on the challenging cases of eye/sensor movements of same direction). It can be verified that the three-point fitted curves can model relatively accurately the eye-tracking behavior of the PSOG subsystem.

To provide a more detailed view of the composite calibration procedure, in Fig. 6 we present the respective low-level (Eq. 9-11) fitted curves that model parameters $a^{H,V}$, $b^{H,V}$, $c^{H,V}$ as a function of sensor position. The overall behavior of the calibration parameters can further justify the universal use of quadratic functions in the developed formulation.

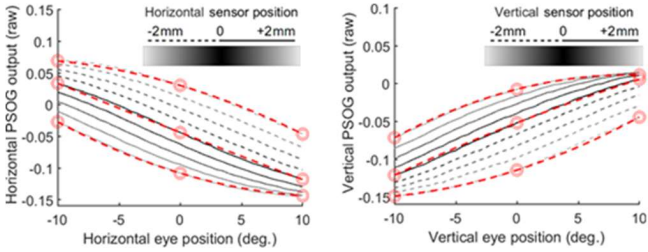

Fig. 5. Curves fitted via the calibration function $f_C^{H,V}$ for the case of eye and sensor movements of same direction.

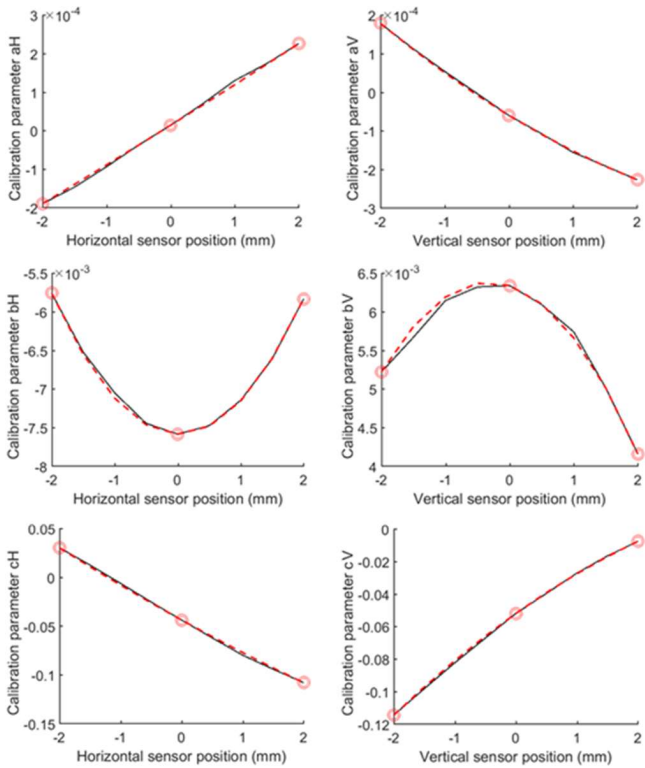

Fig. 6. Curves fitted for parameters $a^{H,V}$, $b^{H,V}$, $c^{H,V}$ for the case of eye and sensor movements of same direction.

### F. Correction Subsystem

The role of the correction subsystem is to process, synchronize, and combine the data streams coming from the PSOG and VOG subsystems in order to generate the corrected output signal. The first step performed by the correction subsystem is to apply any necessary filtering operations to mitigate noise in the raw sample streams. Given the usually low levels of noise of single photodiodes and the fact that we employ a low sampling rate for the VOG subsystem, the existing noise can be smoothed by using low complexity filters that can run in real-time. In our current experiments we use a simple moving average filter of three points (Eq. 12, n = 3) but another attractive option for real-time operation is the Kalman filter. Also, the correction subsystem needs to perform the synchronization of the high sampling rate PSOG data stream and the low sampling rate VOG data stream. This is done by applying zero-order hold filtering on the incoming VOG samples, as presented in Eq. 13 ($f_{IROG}$ = 1000 Hz, $f_{VOG}$ = 5 Hz). After these initial pre-processing steps the correction subsystem is ready to use the sensor movement information in order to select the calibration parameters and perform the fusion of data streams. The calculation of the final sensor-shift corrected signal is performed by applying the inverse calibration function to combine the pre-processed samples from PSOG and VOG subsystems, as shown in Eq. 14.

$$I_{PSOG}^{H,V}{}'(i) = \frac{1}{n} \cdot \sum_{j=0}^{n-1} I_{PSOG}^{H,V}(i-j), i = 0,1,2,\dots,N_{PSOG} \quad (12)$$

$$x_{VOG}^{H,V}(k+\tau) = x_{VO}^{H,V}(k), \tau = 0,1,\dots,\frac{f_{PSOG}}{f_{VOG}} - 1, k = 0,1,2,\dots,N_{VOG} \quad (13)$$

$$y_{PSV}^{H,V} = f_C^{-1\,H,V}\left(I_{PSOG}^{H,V}{}', x_{VOG}^{H,V}{}'\right) \quad (14)$$

The selection of the correct root (from the two) of the inverse quadratic function can be performed by considering the exact domain and range of the original function.

## IV. EVALUATION RESULTS

### A. Experiments

#### 1) Sign conventions for eye and sensor movements

Throughout the paper we use the following sign conventions: horizontal eye movements are positive when the (left) eye moves towards the nasal area, and negative when it moves away from the nasal area. Vertical eye movements are positive when the eye moves downwards, and negative when it moves upwards. Horizontal sensor movements are positive when the sensor moves away from the nasal area, and negative when the sensor moves towards the nasal area. Vertical sensor movements are positive when the sensor moves upwards, and negative when it moves downwards.

#### 2) Eye-tracking scenarios

The experiments for the evaluation of the proposed technique are performed using two eye-tracking scenarios. In both scenarios, real eye-tracking signals are used as input to the simulation framework. The signals were captured from human subjects using an EyeLink 1000 eye-tracker [21] (vendor reported accuracy 0.5°) at a sampling rate of 1000 Hz (monocular setup, left eye was captured). Subjects were positioned at a distance of 550 mm from a computer screen (size 297 x 484 mm, resolution 1050 x 1680 pixels) where visual stimuli were presented. The subjects' heads were restrained using a head-bar with a forehead.

The stimulus of the first eye-tracking scenario (HV) was a 'jumping' point (horizontal/vertical 'jumps') changing its position every 1 second (total duration 36 seconds). The amplitude of the 'jumps' increased from ±2.5° to ±10°, with step of 2.5°. This stimulus-type induces horizontal and vertical saccades of respective amplitudes as the eye moves from one point of fixation to another, and allows for the controlled investigation of the eye-tracking behavior using various eye/sensor movement combinations, e.g. horizontal sensor movements when vertical eye movements occur etc. In Fig. 7 (top), we present the exact ground truth signals for the first eye-tracking scenario (HV). The second eye-tracking scenario (TX) used a text stimulus, in specific, a few lines from the poem of Lewis Carroll "The Hunting of the Snark". We kept a part of the signal corresponding to 10 seconds of reading. The text stimulus allows to explore the eye-tracking behavior in a less constrained scenario with combined horizontal and vertical eye movements. In Fig. 8 (top), we present the exact ground truth signals for the second eye-tracking scenario.

### 3) Sensor movements

For both scenarios (HV, TX) we perform simulated sensor movements by changing the position of the camera module of our setup. Each simulated sensor movement lasts for 2.5 seconds for the TX scenario and 4 seconds for the HV scenario. The magnitudes of the simulated sensor movements cover a range of ±1.75 mm (horizontal/vertical) with step of 0.5 mm. The sensor movements were performed in different parts of the signal, resulting thus in a variety of experimental combinations of eye and sensor movements. To ensure that we cover the most extreme cases for the used ranges of eye and sensor movement we explicitly performed the largest sensor movements (±1.75 mm) at parts of the signal where the largest eye movements occurred (±10°).

### B. Baseline Performance

In this section, we examine the baseline characteristics of the hybrid PS-V technique when assuming no sensor movement (the correction mechanism is inactive). In Fig. 7-8 (HV-TX scenarios) we present the eye-tracking ground truth signals (top), the output signals from the simulated hybrid PS-V technique (middle), and the absolute approximation error between these signals (bottom). For the HV scenario we can see that the approximation error remains at levels under 1° for most of the tested eye movement amplitudes. The observed fluctuations can be attributed to the exact 'goodness-of-fit' of the calibration function at different eye positions. We can also observe the interferences (crosstalk) in horizontal and vertical channels, manifested as apparent small 'saccades' of increasing amplitude in the output of one direction (e.g. horizontal) when eye movement activity occurs on the opposite direction (e.g. vertical). The crosstalk in the vertical output when horizontal eye movements occur appears to be relatively larger. For the TX scenario the horizontal and vertical eye movements follow a more complex pattern, and as a result, the approximation error is dispersed differently than in the previous case. Once again, the error remains at relatively low levels of less than 1°. Since in TX scenario the eye movements can be performed simultaneously in both directions, the observation of crosstalk in this case cannot provide reliable information.

In order to further quantify accuracy and crosstalk we use the measures presented in Eq. 15-18. To use these formulas, first, we need to manually identify the parts of the signals that correspond to fixations, and then, to select the samples from the interior of these parts to avoid outliers during transitions. The accuracy for every single fixation is calculated using Eq. 15-16. For the HV scenario, $N_H$ denotes the number of fixations during the execution of horizontal saccades (seconds 1-17) and $N_V$ denotes the number of fixations during the execution of vertical saccades (seconds 18-34). For the TX scenario, where combined movements are performed, it stands $N_H = N_V$ (seconds 1-10). In all cases $M$ is the number of samples within the fixation under consideration ($M$ changes from fixation to fixation). The output samples for the hybrid

PS-V technique are denoted as $y_{PSVj}^{H,V}$ and the ground truth samples are denoted as $y_{Gtj}^{H,V}$. For the calculation of crosstalk we use Eq. 17-18, with $cross_i^{HV}$ denoting the crosstalk in horizontal channel when vertical eye movements occur, and $cross_i^{VH}$ denoting the crosstalk in vertical channel when horizontal eye movements occur. As we mentioned before, the calculation of crosstalk for the TX scenario does not provide reliable information, and for this reason, crosstalk is quantified only for the HV scenario. In Table 1, we present the summarizing results for accuracy and crosstalk showing the calculated mean values (M) and standard deviations (SD) over all respective fixations in each case. We can observe slightly different behavior of horizontal and vertical accuracy in HV and TX scenarios, which can be partially attributed to the fact that in TX scenario the maximum absolute horizontal eye movement range is relatively smaller than in HV scenario.

$$acc_i^H = \sum_{j=1}^{M} \left| y_{PSVj}^H - y_{Gtj}^H \right| / M, \, i = 1, \ldots, N_H \quad (15)$$

$$acc_i^V = \sum_{j=1}^{M} \left| y_{PSVj}^V - y_{Gtj}^V \right| / M, \, i = 1, \ldots, N_V \quad (16)$$

$$cross_i^{HV} = \left| \frac{\sum_{j=1}^{M} y_{PSVj}^H - \sum_{j=1}^{M} y_{Gtj}^H}{\sum_{j=1}^{M} y_{Gtj}^V} \right|, \, i = 1, \ldots, N_V \quad (17)$$

$$cross_i^{VH} = \left| \frac{\sum_{j=1}^{M} y_{PSVj}^V - \sum_{j=1}^{M} y_{Gtj}^V}{\sum_{j=1}^{M} y_{Gtj}^H} \right|, \, i = 1, \ldots, N_H \quad (18)$$

TABLE I
Baseline accuracy and crosstalk for the hybrid PS-V technique

| | Accuracy (°) | | | Crosstalk (%) | |
| --- | --- | --- | --- | --- | --- |
| | **H** | **V** | | **H-V** | **V-H** |
| **Scen.** | M (SD) | M (SD) | **Scen.** | M (SD) | M (SD) |
| **HV** | 0.38 (0.11) | 0.30 (0.12) | **HV** | 4.7 (1.9) | 5.8 (5.3) |
| **TX** | 0.28 (0.15) | 0.39 (0.17) | | | |

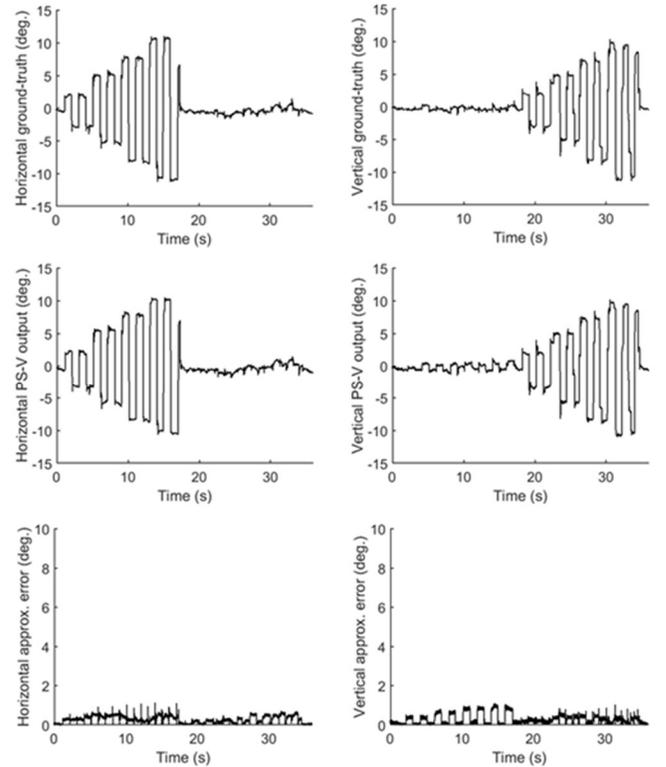

Fig. 7. Eye-tracking scenario HV: ground truth signal (top), hybrid PS-V output (middle), approximation error (bottom).

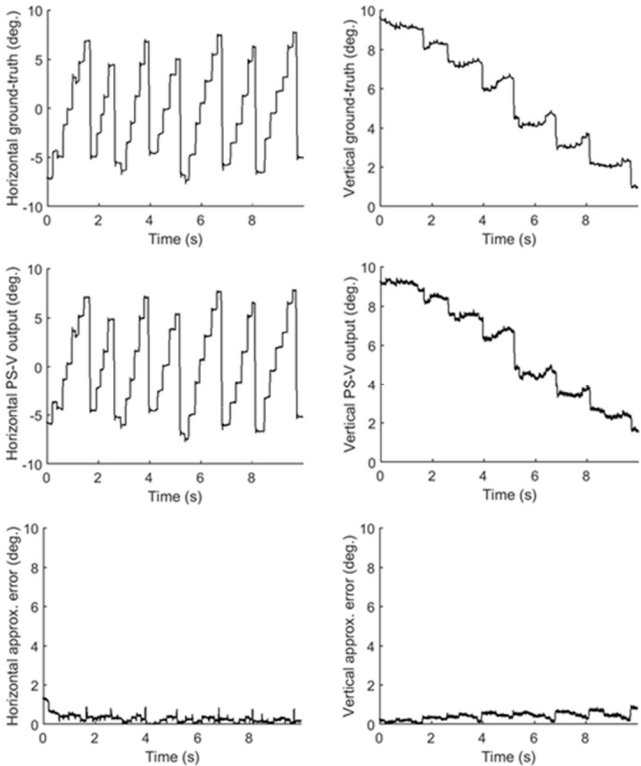

Fig. 8. Eye-tracking scenario TX: ground truth signal (top), hybrid PS-V output (middle), approximation error (bottom).

## C. Performance for Sensor Movements

In the event of sensor movements, the traditional PSOG techniques cannot cope with the translation of the captured areas and can no longer perform with accuracy the transformation from raw output units to degrees of visual angle. This can cause shift-induced deformations on the output signal. In Fig. 9 (top) we show the ground-truth signal for the HV scenario (during horizontal saccades) and in Fig. 9 (bottom-left) the resulting output when using traditional PSOG (without correction), for the case of an example sequence of sensor movements. In Fig. 9 (bottom-right) we show the resulting output when using the hybrid PS-V technique.

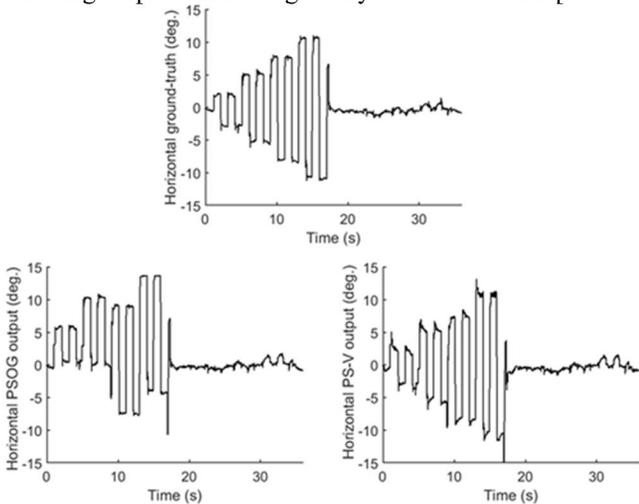

Fig. 9. Sensor-shift induced artifacts: the ground truth signal (top), the deformed output signal from traditional PSOG (bottom-left), and the corrected output signal from the hybrid PS-V technique (bottom-right).

We can qualitatively observe that due to the novel scheme that uses the estimated sensor movement to correct the eye-tracking signal, the hybrid PS-V technique can lead to substantial decrease of the sensor shift-induced deformations. In next sections we present the quantitative results of the performance of the hybrid PS-V technique when sensor shifts occur.

### 1) Sensor movement estimation error

Our evaluation experiments involved the execution of simulated sensor movements in range ±1.75 mm at different parts of the signals from HV and TX scenarios. Before we examine the final eye-tracking accuracy it would be valuable to inspect the error of the sensor movement estimation process itself. In Fig. 10, we present diagrams that show the error in the estimated sensor movement compared to the ground truth. The diagrams correspond to the aggregated results from HV and TX scenarios. The points represent the calculated mean values over all samples that correspond to each performed sensor movement, and the error-bars show the respective standard deviations. The theoretically perfect estimation is denoted by a dashed line. We can observe that the deviations of the estimated points from the line of perfect estimation are in most cases within the range of ±0.2°. Also, we can see that the vertical movement estimation seems to be slightly less accurate at the extremes of the tested range. This can be attributed to the wide capturing angle that can disrupt capturing of pupil center and corneal reflection(s) at these positions.

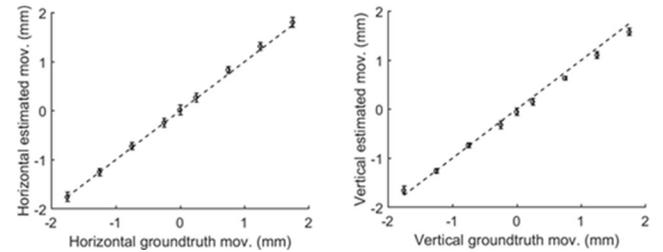

Fig. 10. Sensor movement estimation error (mean, standard deviation). Perfect estimation denoted with dashed line.

### 2) Evaluation of eye-tracking accuracy correction

In order to evaluate the afforded eye-tracking accuracy correction when sensor movements occur we will employ the formulas of Eq. 15-16. In this case, though, the used fixations are the ones from the signal parts where the corresponding sensor movements occurred. In Fig. 11, we present the eye-tracking accuracy diagrams comparing traditional PSOG and the hybrid PS-V technique, for the HV scenario. We can clearly observe the large intolerance of the PSOG technology to sensor movements, especially when these movements occur at the same direction with the eye movements (Fig. 11 top-left, bottom-right). For sensor movements of ±1 mm the accuracy error can go at levels of over 5°. On the other hand, for the hybrid PS-V technique we can observe only a small increase of the accuracy error when increasing the magnitude of sensor movements, with the error generally kept at levels under 1°. For the case of different directions of sensor and eye movements the used PSOG design appears to be also relatively robust, however, even in this case the hybrid PS-V technique is more consistent for the whole range of sensor movements. An interesting observation is that in both cases the eye-tracking accuracy diagrams are not totally symmetric for positive and negative sensor movements. This should be attributed partially to the asymmetries of the eye, e.g. shape, upper and lower eyelids, and partially to the exact used experimental combinations for the relative magnitudes of eye and sensor movements. It should be emphasized that the exact same experimental combinations were used both for traditional PSOG and for the hybrid PS-V technique in order to ensure the clear investigation of the achieved improvements in accuracy when using the developed technique. In Fig. 12, we present the corresponding diagrams for the TX scenario. The overall trends are similar as for the HV scenario, however, the horizontal eye-tracking accuracy error for traditional PSOG seems to rise even more steeply, whereas for the hybrid PS-V technique the respective error appears to be slightly larger and more variable than previously, possibly affected by the combined execution of eye movements at both directions.

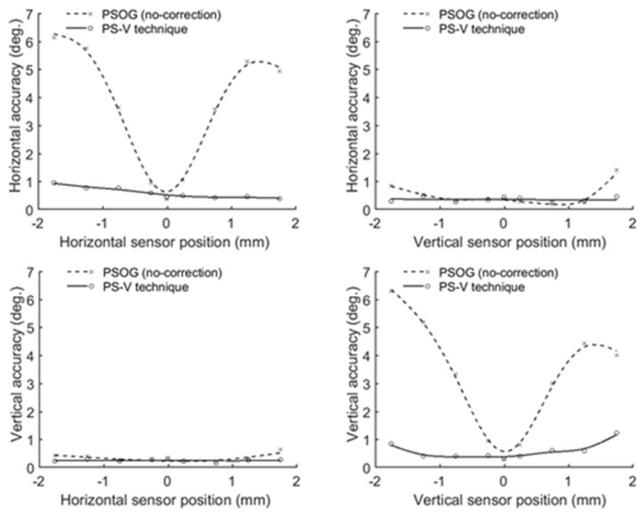

Fig. 11. Eye-tracking scenario HV: accuracy vs. sensor movement for traditional PSOG and the hybrid PS-V technique.

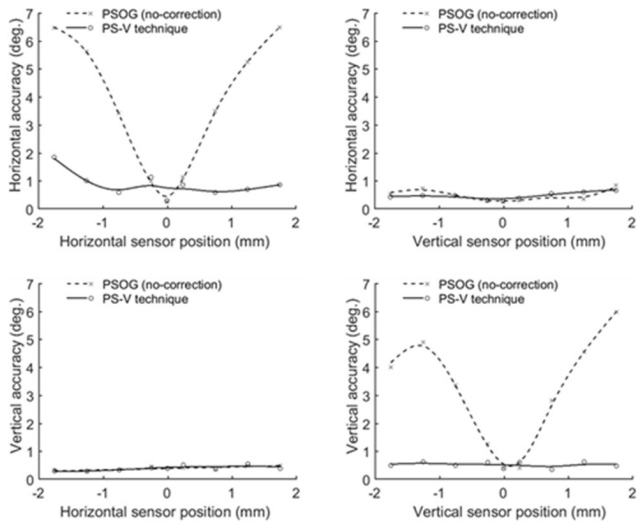

Fig. 12. Eye-tracking scenario TX: accuracy vs. sensor movement for traditional PSOG and the hybrid PS-V technique.

### 3) Demonstration of correction in a practical application

To demonstrate the practical importance of the afforded correction we present an example of the effects from using traditional PSOG and the hybrid PS-V technique in a simplified application of foveated rendering [1]. In this example, we utilize data from the TX eye-tracking scenario and we demonstrate the effects from sensor movement on the selection of the 'foveated rendering area', i.e. the region around the user's point of gaze that will be rendered with high quality (in this case higher resolution). In Fig. 13 on the left column we present the ground truth eye-tracking signal (top), the traditional PSOG output (middle), and the hybrid PS-V technique output (bottom), for the case of a vertical downward sensor movement of 1.25 mm occurring between seconds 5-7 during the TX scenario (only the vertical component of movement is shown). On the right column of Fig. 13 we show for a specific point in time (within the duration of sensor shift) the respective gaze points and the areas that are rendered with high-resolution on the original text stimulus. The gaze points are back-projected on the stimulus space using the signals and the exact parameters of our experimental setup. The rendered areas are modeled as circles with center the respective gaze point each time and radius of 5°, chosen to resemble the size of the central fovea of the eye [1]. To facilitate inspection, the gaze point from the ground truth signal (where the user really looked at) is shown with a cross and the gaze points estimated with either of the two methods are shown with small circles.

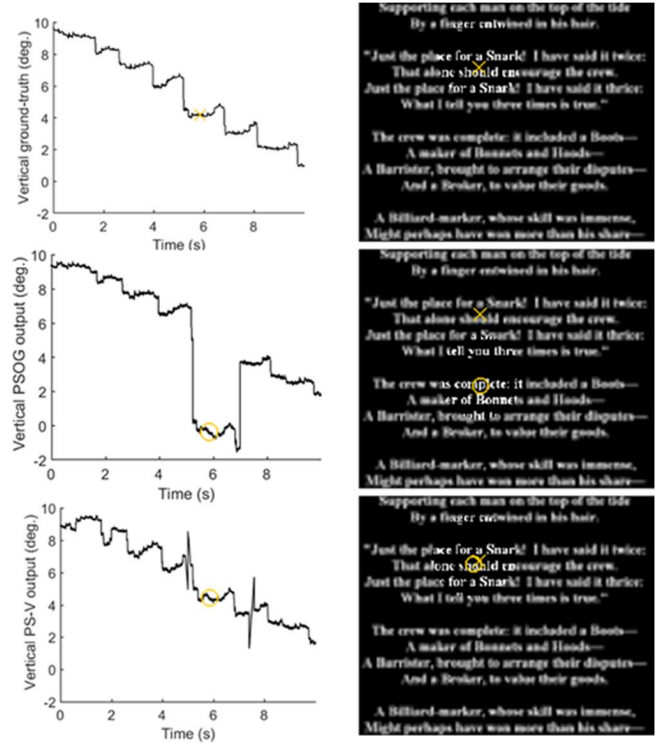

Fig. 13. Demonstration of eye-tracking signals and corresponding effects on the high-resolution rendering area for a vertical sensor movement of +1.25 m.

As we can observe, when using traditional PSOG without correction the high-resolution rendering area is far away from the really attended area, thus generating an unacceptable effect for the user. On the other hand, when using the hybrid PS-V technique we can see that the achieved correction, which brings accuracy error at levels lower than 1°, results on the high-resolution rendering of an area that encloses the attended location at a sufficient degree. Practically, this means that the high-resolution rendering area can be expanded with minimal computational cost in order to make the error less perceptible to the user. It should be noticed that the brief spikes that appear in the corrected signal are an expected effect of the non-zero duration of the sensor shift transition phases in combination to the low sampling rate of the VOG correction mechanism (see discussion about delay in next Section).

## V. DISCUSSION

### A. General Analysis of the Hybrid PS-V Technique

The main goal of our current experiments was to assess the degree at which the developed hybrid PS-V technique is capable to address the inherent flaws of traditional PSOG technology when sensor movements occur. Our results show that the afforded improvements in eye-tracking accuracy are considerable, and more importantly, the accuracy error due to sensor movements is kept at levels of 1° or lower, which can be acceptable for various HCI applications. The investigated performance covers a window of eye movements in range of ±10° and sensor movements in range of ±2mm. As it can be observed in Fig. 4, though, for sensor movements smaller than 1 mm the output from the PSOG subsystem remains relatively linear even for a larger range of eye movements (e.g. up to ±15°). This means that depending of the requirements of the application under consideration, the technique allows the possibility to give more weight on the coverage of larger eye movement range at the expense of covered sensor movement range and vice versa. The developed simulation framework gave us the opportunity to experiment with different parameters for the used PSOG design, such as the size, positioning, and overlap of the simulated sensor reception areas. Our experimentation showed that the variation of these parameters can affect conversely the properties of linearity and crosstalk of the system. The final specifications for the used PSOG design were selected with the aim to provide a good trade-off between these characteristics for the target ranges of eye movements and sensor movements. It should be noticed that although further algorithmic optimizations might be possible, it is expected that additional hardware (e.g. more IR sensors) will be required for the simultaneous coverage of much larger range of eye and sensor movements. Since from a point and after the sensors move to areas where no useful information can be captured, the problem is no longer of algorithmic/processing nature but lack of information.

Regarding the sensor movement estimation via the VOG subsystem, the investigation of Fig. 10 reveals the expected levels of error when detecting the pupil center and corneal reflection under various conditions. The observed inaccuracies expose some inherent limitations of the pupil center-corneal reflection technique, since the apparent movements of these features can be affected by factors such as the distance and the viewing angle of the capturing sensor. Although such inaccuracies are overshadowed by the levels of afforded error correction for larger sensor movements, for the case of small or no sensor movements the resulting artifacts can become more prominent. For example, as we can see in Fig. 13, although the correction for the part of the signal when the sensor shift occurs is remarkable, the rest of the signal appears to have small deformations when compared to the ground truth and the uncorrected signals (similar small deformations can be seen in Fig. 9). A possible method to mitigate such small deformations during a practical implementation is to use a hard (or adaptive) threshold in order to activate/deactivate the correction mechanism when the estimated levels of sensor movement are above/below the selected threshold.

An interesting prospect for the current technique is the possibility to perform the calibration for sensor movements in an automatic (and thus more user-friendly) manner. This option can be performed by using the VOG subsystem during the calibration process as a feedback mechanism to indicate sensor position. Practically, this means that during calibration the user will not be needed to accurately place the sensor at pre-defined points but he/she can move the sensor arbitrarily (but within the limits of the desired range) and allow the VOG subsystem to feed the $x_s$ values needed for Eq. 9-11. It should be emphasized that for such an operation the reliability of the movement estimation algorithm is of utmost importance because any errors at this early stage will be propagated and affect the overall correction performance. To test the automatic calibration method for the current setup we performed additional experiments, and in Fig. 14 we present the resulting curves (dot-lines) when performing calibration via feedback from the VOG subsystem. During these experiments we moved the sensor at positions -2, 0, and +2 mm and then instead of using directly these values during calibration we used the values estimated by the VOG subsystem. The fitted curves are superimposed on the original curves presented in Fig. 5 (where the exact sensor positions were used). As we can observe, the fitted curves for the horizontal sensor movements are very close to the original calibration curves whereas for the vertical sensor movements there is more noticeable deviation at the extremes. This behavior reflects the slightly less accurate sensor movement estimation for the vertical sensor movements, previously observed in Fig. 10.

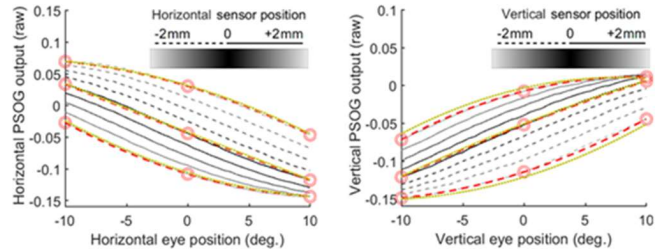

Fig. 14. Demonstration of fitted curves for the automatic sensor movement calibration process with feedback from the VOG subsystem.

Another important aspect when considering the practical application of the hybrid PS-V technique regards the delay needed to detect sensor movements. This delay combined with the non-zero duration of the sensor shift transition phases can result in brief artifacts (spikes) in the signal, as shown in Fig. 13. Given that the sensor movement estimation is performed by the VOG subsystem, it is the low sampling rate of this subsystem that mainly sets the bounds for the maximum expected delay. As a result, for the current setup with the VOG subsystem running at 5 Hz this maximum delay is expected to be about 200 ms. This delay can be acceptable when considering the expected frequencies-of-occurrence and durations of events that usually induce sensor shifts in head-mounted devices, like facial expressions and body movements. It should be clarified that this delay is related only to the periods when sensor movements occur since for the rest period of normal operation the system tracks with the fast rate warranted by the low sampling interval (1000 Hz, sampling interval: 1 ms) and the low computational complexity of the PSOG subsystem.

### B. Practical Considerations for Power Consumption, Computational Complexity and Cost

The hybrid PS-V technique can potentially provide significant gains when compared to systems based on VOG in terms of power consumption when fast eye-tracking is required. Commercial high-speed eye-tracking systems based on VOG (e.g. EyeLink 1000 [21]) can consume power of several Watts, and their demands can be fulfilled only via tethered operation. At a research level, there have been some recent efforts focusing to push the limits of VOG power consumption under 100 mW [22, 23], however, such optimizations can impose certain limitations on operational accuracy and sampling rate.

During high-speed operation of a VOG system there are two main sources that can inflate power consumption rather sharply: pixel acquisition and increased computational burden of image processing. The hybrid PS-V technique performs the high-speed acquisition part with the PSOG subsystem that is based on simple IR sensors. This gives the ability to operate at high sampling rates while keeping the required increase in power at minimum levels. Typical IR sensors can have total power dissipation of 100-200 mW, however, due to their fast switching times (order of tenths of ns) duty-cycle control can be applied and combined with voltage optimizations can result in power consumption at the order of tenths or hundredths of μW [23]. Considering now that the PSOG subsystem uses just a few IR sensors (in contrast to thousands of pixels) the power consumption for the high-speed (e.g. 1000 Hz) acquisition of samples from the PSOG subsystem can be expected to be less than 1 mW. The hybrid PS-V technique uses also a VOG subsystem but this subsystem operates constantly at low sampling rate since it is used only for correction when sensor movements occur. This allows to keep the power requirements for the VOG subsystem at a minimum level. The second source of power efficiency of the hybrid PS-V technique is the low computational complexity. The current PSOG design requires only four additions and two subtractions for the combination of the sensor single-valued outputs. Just a few more simple operations are needed for applying the calibration mapping function (calibration parameters are pre-calculated) and the running average filters. Hence, the total number of operations will be just a tiny fraction of the operations needed by a pure VOG system operating at a high rate (thousands or even millions operations per frame). As previously, the VOG subsystem of the hybrid PS-V technique produces a steady overhead irrespectively of any increase of the eye-tracking rate that is governed by the low-complexity PSOG subsystem. Based on the discussed considerations the total power consumption (acquisition and processing) of a system based on the proposed hybrid PS-V technique is expected to be under 15 mW while operating at high sampling rates of 1000 Hz or more.

Another important advantage of the hybrid PS-V technique from a system's perspective is the ability to keep the overall cost at very low levels when compared to a pure VOG implementation running at 1000 Hz. Once again, the reason is that in the hybrid PS-V technique the demanding high-speed eye-tracking part is achieved via the PSOG subsystem. The cost of the IR photo-sensors and a typical video camera can be kept at the order of tenths of dollars, whereas on the other hand the cost of a video camera operating at 1000 Hz can be hundreds of dollars. The large difference in the required budget becomes even more prominent when considering binocular eye-tracking, which is expected to be the norm for emerging interaction devices like the AR/VR headsets.

### C. Current Limitations and Future Extensions

The current evaluation of the proposed hybrid PS-V technique was done within the scope of certain limitations. The calibration function is trained with eye and sensor movements performed separately at the horizontal and vertical directions. The proposed technique can be further strengthened by exploring more generalized calibration functions suitable to cope with scenarios involving (large) oblique eye movements, and also, rotational and depth sensor movements. Furthermore, the currently used sensor movement estimation algorithm assumes that at least one corneal reflection can be traced on the eye image. However, such an assumption can pose certain limitations for the positioning of sensors. The investigation of alternative mechanisms for sensor movement estimation based on other characteristics (e.g. pupil-ellipse shape) can provide more flexibility when larger eye movement ranges and/or wider sensor positioning angles are required. Even though we described the favorable characteristics of the hybrid PS-V technique in terms of power consumption, the limits can be pushed even further with the development of a low-complexity mechanism that can detect reliably the onsets and offsets of sensor movements. This mechanism would allow operating the VOG subsystem at an asynchronous 'detection-triggered' rate, and also, it would assist on the mitigation of spike-artifacts appearing during the transition periods. Last, the hardware implementation of the technique on a head-mounted device (e.g. an AR/VR enabled headset) can allow for the exploration of possible design improvements and the detailed examination of real-eye artifacts that might not be covered sufficiently by the current simulation.

## VI. CONCLUSION

In this paper we described the hybrid PS-V technique, a novel approach that combines the principles of photosensor and video oculography in order to tackle the accuracy issues of traditional photosensor oculography when sensor shifts occur. Our investigation was based on the use of a semi-simulated framework making possible to explore the behavior of different components of the technique in a controlled manner, and leading to the formulation of a composite calibration model that can be used to effectively combine the information coming from the PSOG and VOG subsystems. The results from our evaluation experiments demonstrate the large accuracy improvements that can be achieved for sensor movements in range of ±2 mm. The achieved levels of correction combined to the favorable characteristics of the photosensor oculography subsystem, reveal the promising prospects for using the hybrid PS-V technique to enable high-speed, low-power eye-tracking in modern head-mounted devices.